# RAPID DEVELOPMENT OF A MOBILE ROBOT SIMULATION ENVIRONMENT

Gordon Stein,[*] CJ ChanJin Chung,[†]

Robotics simulation provides many advantages during the development of an intelligent ground vehicle (IGV) such as testing the software components in varying scenarios without requiring a complete physical robot. This paper discusses a 3D simulation environment created using rapid application development and the Unity game engine to enable testing during a mobile robotics competition.

Our experience shows that the simulation environment contributed greatly to the development of software for the competition. The simulator also contributed to the hardware development of the robot.

**INTRODUCTION**

Simulations have been a major part of robotics research and development for decades. Simulation allows for navigation concepts to be tested without requiring access to a robot or the physical environment where the robot is intended for use.

This paper reports on the creation of a robotics simulation environment using the Unity game engine. Existing robotics simulations often have issues with cross-platform compatibility, graphics fidelity, and ease of modification. Many existing simulators, such as Gazebo[6], are only available on Linux or have poor support on the more common Windows and Mac OS platforms. Other simulators, such as Stage[7], only simulate a two-dimensional world, making them less useful for robots navigating through vision. Proprietary simulators are also available[8], providing the desired features, but at a possibly restrictive combination of price and licensing.

Game engines have been developed with a focus on the same features required of a robotics simulation platform, making a game engine an ideal starting point for creating a new robotics simulation environment. Unity provides a free development environment to create content for 22 platforms, including Windows, Linux, Mac OS, Android, and iOS[2]. In addition, Unity has previously been used for robot design and simulation to a limited extent[4], and another game engine, Unreal Engine, was used to develop USARSim.[9]

In Section 2 of this paper, the goals of the simulator are explained, along with a description of the physical robot being simulated and the software used. Section 3 describes the implemented functionality of the simulator. Our experience of using the simulation environment and future work are discussed in later sections.

---

[*] Graduate Student, Department of Math and Computer Science, Lawrence Technological University, 21000 W. Ten Mile Road, Southfield, MI 48075-1058, USA.
[†] Full Professor, Department of Math and Computer Science, Lawrence Technological University, 21000 W. Ten Mile Road, Southfield, MI 48075-1058, USA.

**SIMULATION ENVIRONMENT DESIGN**

**Simulation Environment Goals**

The simulation environment was created to accelerate development of a robot for the Intelligent Ground Vehicle Competition (IGVC). There are three competitions within the IGVC: the autonomous navigation (auto-nav) challenge, the design competition, and the interoperability profiles (IOP) challenge[1]. The simulation would be most relevant to the auto-nav and IOP challenges.

The auto-nav competition requires the robot to navigate an obstacle course consisting largely of construction barrels placed in a lane marked by white lines painted on grass. The simulation environment would allow the team to test its code for detecting these obstacles and navigating around them without requiring a course be set up. It would also allow testing to continue during inclement weather, when testing the robot outside might not be possible, during night, when the lighting conditions no longer reflect the conditions at the real course, and while the physical robot was under construction. In addition, the simulation environment could be tuned to the conditions of the course at the time of the competition. Seasonal change prevented a physical course from accurately representing the lighting and grass conditions present on the official course during the early June weekend of the competition.

In the IOP challenge, a computer is provided to act as a common operating picture (COP). The robot's computer must communicate with the COP using the JAUS standard[10]. There are a series of tasks the COP will send the robot, and the robot is scored on its responses. Several of these tasks require the robot to follow instructions in a large outdoor area. The simulation environment allows the robot's JAUS code to be tested without needing to have the robot outside.

A level of realism was desired from the simulation environment so that the results of the simulated robot would be useful when the same code is run on the physical robot. This consists of several goals:

- The simulated motor controller responds to commands in the same manner
- The robot's motion after being given a command is in the correct direction and magnitude for both forward and angular velocities
- Obstacles present in the simulation reflect likely obstacles present on the IGVC auto-nav course
- Input from simulated sensors reflects how the sensors would behave in the same situation

The simulation environment was created using rapid application development principles. This development method "…uses small teams, fourth generation tools, fast development, tight timescales and resource constraints, iterative, evolutionary and participative prototyping and intensive user involvement amongst a variety of principles."[3] With the limited time between the beginning of development and when the team needed to begin testing, rapid development was required. As the simulation environment was developed, feedback from the team working on the navigation software was used to improve it.

**Robot Simulated**

The robot simulated in the software was "Bigfoot" (Fig. 1), Lawrence Techno-logical University's 2015 IGVC entry. Bigfoot was built on top of a Clearpath Husky chassis, with additional hardware to add a computer and sensors. These sensors included a Microsoft Lifecam camera, a Hokuyo URG-04LX-UG01 LIDAR, a Sparton GEDC-6E digital compass, and a Novatel Power-Pak differential GPS unit.

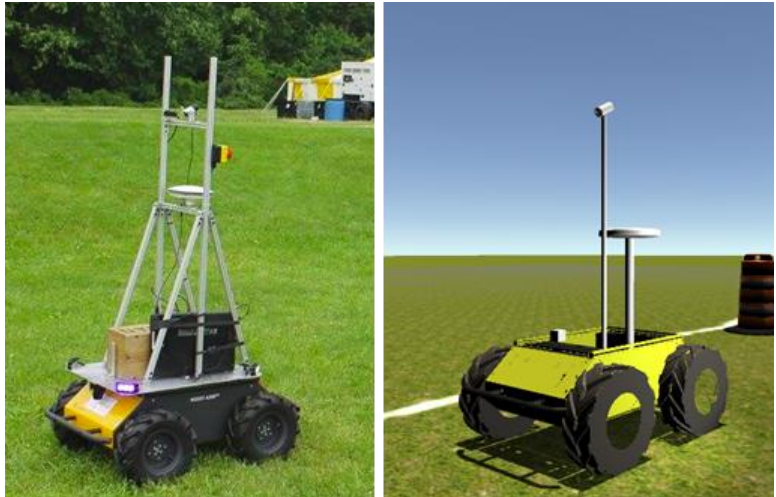

**Figure 1    Physical and Simulated Robot**

The Husky is a simple robot to simulate due to the operation of its motor controller and the simplicity of its drive systems. Commands are sent to the motor controller by providing it with desired forward and angular speeds, in meters per second and degrees per second respectively. The internal motor controller on the Husky chassis reads information from the encoders to determine the correct power to provide the motors to reach the requested speed. The motors on the Husky create "skid-steer" locomotion, allowing the robot to turn in place. These factors allow the simulator to be simplified by setting the simulated robot's forward and angular velocities to approach the received desired values instead of creating a complex wheel dynamics simulation. The Husky chassis also lacks any significant suspension system, so the robot's body can be simulated as a single rigid body with wheels directly attached to it.

**Software Used**

Unity, a game engine from Unity Technologies2, was used to create the simulation environment. Unity was chosen for its ease of use and that additions to it can be created using the C# programming language, preventing developer retraining, and allowing a potential for code reuse with the robot's navigation code. This software supports the rapid application development strategy by allowing for changes to easily be made as the simulation environment is created. Unity also includes easy to use tools for model importing and terrain creation, allowing the virtual course to be set up without a large time requirement.

The robot navigation software was developed using Microsoft Visual Studio and the C# programming language. This language was chosen based on its flexibility and the existing knowledge of the team members. The code used for the simulator was also written in C#, but using the MonoDevelop IDE.

## IMPLEMENTATION

### Environment Design

A model of Bigfoot (Figure 1) was created for use in the simulation environment. The main geometry of the model was provided by Clearpath. The provided model had unnecessary details such as the interior of the robot chassis, which were removed to improve the performance of the simulation environment.

The simulated model of Bigfoot does not exactly match the physical version, as it was created before the physical robot was built or fully designed itself. However, the components that are not present in the simulation model, such as the support structure, are not functional in the simulator and are not visible to the robot's sensors.

All of the robot's sensors are simulated. The simulated camera is created through a camera object in Unity, set to render to a texture. The simulated versions of the GPS and compass use determine the represented latitude, longitude, and heading of the robot based on its position in the simulated space. The LIDAR uses Unity's raycasting features to test the same angles and distance that the physical LIDAR scans over. Each sensor has a class creating a Unity component, each providing a common interface for retrieving the sensor output.

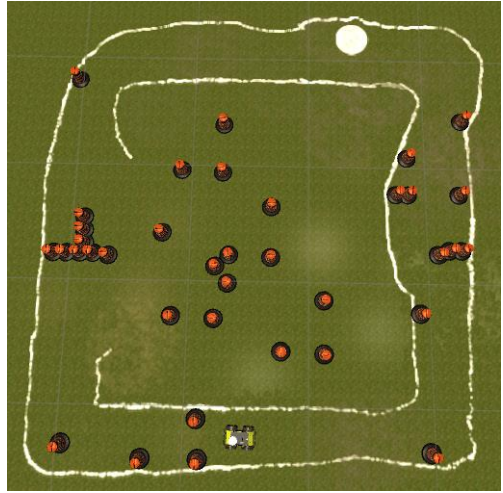

**Figure 2    Overview of the Simulated IGVC Course**

A simulated course based on the 2014 IGVC auto-nav course (Fig. 2) was created for the simulated robot to navigate through. This course featured the construction barrels and white lines from the course, along with several other obstacles that could possibly appear on the 2015 competition course. Unity's terrain painting feature was used to create variation in the color of the grass and line. The simplicity of the Unity editor allowed for the field to be created in under an hour.

### Communications

The ConnectionManager class manages communications between the simulation environment and connected software components (Figure 3). Ports for each sensor are given to an instance of this class. ConnectionManager will establish a TCP connection on each sensor port on the local machine upon request. Once the connections are set up, the ConnectionManager polls each of the sensors when the simulation environment updates, and sends the new data if the sensor object reports that it has updated since the last polling.

A similar client-server architecture is present in other available robotics simulators.[7] By providing a simple interface to communicate, the simulator can potentially communicate with many different robotics platforms. In addition, the communications could be sent to a client on a different computer, allowing the robot's actual computer hardware to be used, without the additional computational strain of running the simulation environment at the same time as the robot's navigation code.

UDP was also considered for the communications method, but TCP was chosen instead for multiple reasons. While UDP would allow for datagrams to be passed with less overhead, the camera data would need to be broken up into multiple smaller portions and reassembled in the robot code. TCP allows the data to be sent as a single stream without a need for the software on both ends to perform any additional operations. Other methods of inter-process communication, such as shared memory and pipes were considered, but rejected due to the added dependencies required for platform independence, and the inability to easily use these methods over a network for future work.

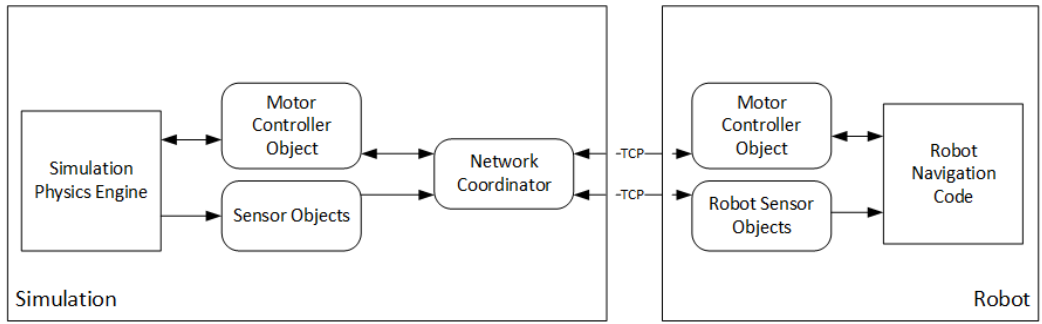

**Figure 3    Communications Between Simulator and Robot Software**

**Integration into Existing Code**

The code for the robot was created with the intent that it would be capable of switching between simulation and physical modes at runtime, without any modifications required to the navigation logic. The navigation code contains references to the Sensor and MotorController superclasses, rather than specific sub-classes. Each class provides the general capabilities a sensor or motor controller will need, which were implemented for both physical and simulated components (see Figure 4).

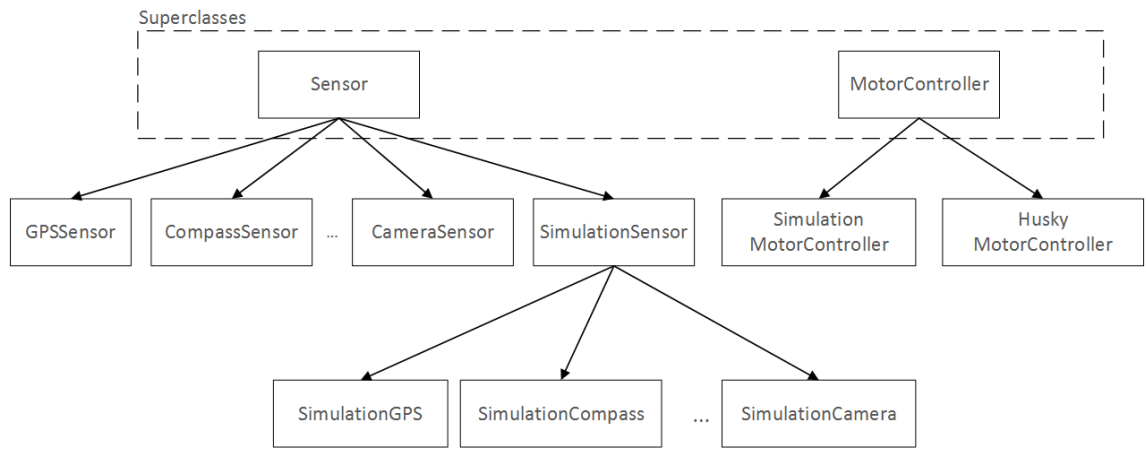

**Figure 4    Class Hierarchy of Sensors and Motor Controllers**

Creating a sensor class to collect data output by a sensor in the simulation environment is a simple process. The class is created as a subclass of the SimulationSensor class. The SimulationSensor class is in charge of setting up the communications between the simulation environment and the robot code, exposing a method to be overridden by subclasses, for processing the incoming datagrams. The Sensor classes were abstracted to provide all communications setup through an initialization method and any communications cleanup through an implementation of the IDisposable interface, so the navigation controller would not need additional code to be switched to using the simulation classes.

For example, the latitude and longitude data from the simulation environment is received by the SimulationGPS sensor class using the following:

```
class SimulationGPS : SimulationSensor
{
        protected override void messageReceived(byte[] message)
        {
           try
           {
               sensorData = new Tuple<float, float> (BitConverter.ToSingle(message, 0), BitConverter.ToSingle(message, sizeof(float)));
           }
           catch (Exception)
           {
               Debug.WriteLine("Error in {0} (most likely due to sudden simulation disconnect)", this.GetType().Name);
           }
        }
}
```

Other sensors require approximately the same amount of code. By creating the SimulationSensor class, the development of new software components that communicate with the simulation environment was greatly simplified.

**Using the Simulator**

Use of the simulator with the Bigfoot code is straightforward. The software created for Bigfoot features a simple interface for switching to the simulation mode. The robot software is initially in neither mode. To select the simulation mode, the "Simulation Mode" button (see Figure 6) must be pressed. This commands the robot software to begin listening for TCP connections from the simulator for each of its sensors and its motor controller. The simulator must then be launched, if it is not already running. Finally, the "Connect" button in the simulator (see Figure 5) must be pressed. This commands the simulator software to open TCP sockets to connect to the sockets in the robot process already awaiting the connection.

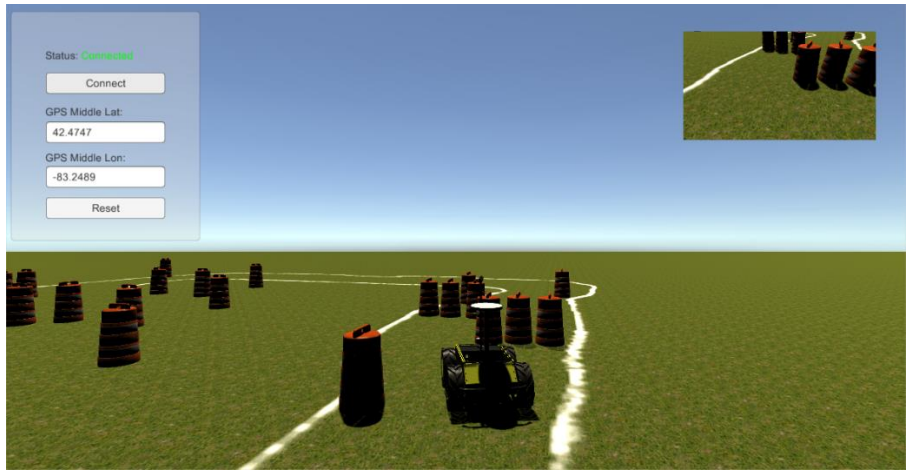

**Figure 5      Screenshot of Simulator Running and Connected to Robot Software**

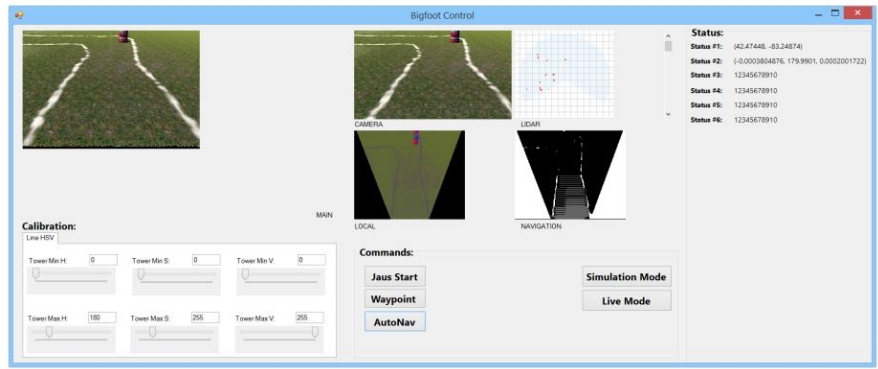

**Figure 6      Screenshot of the Robot Software Connected to Simulator**

**RESULTS**

During the early stages of the robot's development, the simulation environment proved to be very useful. The physical robot was not completed until April 2015, and weather conditions in the months leading up to the competition were rarely suitable for outdoors testing. The simulator was used to test the software architecture and demonstrate the ability of the robot's software to switch between motor controllers and navigation controllers at runtime. Navigation code was written using the simulation environment, allowing the team to make progress before the robot was complete. An advantage of using a simulation is that it pro-vides a controlled testing area for multiple users to work in at the same time. This allows it to be easily used to compare different concepts and algorithms for navigation[5].

The simulator also helped accelerate the design of the physical robot by allowing for simple testing of potential camera angles and heights. The visual editor in the Unity software allowed for new camera positions and orientations to be tested and evaluated before the hardware camera component was mounted to the robot. In previous years, the team arbitrarily chose a camera height and angle and did not change the positioning due to the time required to relocate and test the camera. The simulation allowed for a better decision to be made about the location and orientation of the camera.

The ability to easily change the design of the simulated course was a large advantage at two points. Early in the robot's development, the team wanted to be able to test its ability to avoid the white lines of the course, but not the barrels. This change was easily made using the Unity editor. At the competition, it was observed that the barrels were being set up in a formation that had not been tested by the team. However, creating this obstacle was also easily accomplished because of the Unity engine underlying the simulation environment.

Once the real robot was complete and it was possible to test outdoors, two issues in the simulator became apparent. The main issue was that the color balancing of the simulated camera and the physical camera did not match. This resulted in different calibration settings being required for the simulated and physical cameras. Simulations used for robotic vision commonly have similar issues when attempting to create models for camera systems[5]. The other issue found was that the simulated GPS's coordinate system was scaled incorrectly, but that issue was resolved for the competition.

The simulation environment remained useful for the IOP competition. Development of code for both the auto-nav and IOP competitions was ongoing at the same time during the competition. However, competing in either competition or testing the code for either module requires the robot to be present. To resolve this issue, the simulation environment was used to test the IOP code. The issue with the camera would not have an effect on the IOP code, which does not interact with it, so the code could be quickly tested using the simulation before deploying it to the physical robot and having the robot judged.

**FUTURE WORK**

Future plans focus on redesigning the simulation environment to be easier to configure. The current version, designed only for the IGVC team, requires modifications to be made directly in the Unity editor, and only has the specific sensors required to simulate Bigfoot. Future versions will be created with the intent that they can be used for many different robot platforms and easily modified for new sensors. Unity is capable of loading assets and code at runtime, allowing users to implement novel sensors without having to recompile the entire project. Using Unity also allows for a visual editor for new robots, environments, and components for the simulation environment, without requiring additional development time.

The current sensors included in the simulation environment are both too accurate and not accurate enough. The current implementations of the digital compass and GPS sensors do not include inaccuracies observed with the physical versions. The simulated camera provides very clear images, but they do not include specular effects from bright sunlight. Additional components are available for Unity to provide a more accurate camera model.

One future extension of this work is to create a simulation model of a simple robot and compare its reaction to the real robot. Odometry for both the physical and simulated robots would be used to determine the differences in paths traveled by each of them. This would help to provide empirical validation of the simulation model used.

Another future extension of the simulator is to run it over a network. With the simulation being run on a separate computer on the same local area network as the computer running the navigation code, the resource utilization will be more realistic. Running both the navigation software for the robot and the simulation software on the same computer creates competition for system resources. The performance of the simulator when run over a network could also be analyzed to determine the cost of the overhead involved and the potential for hosting the simulation remotely and providing it as a service.

## ACKNOWLEDGMENTS

The authors would like to thank other IGVC 2015 team members: Fan Wei, Icaro Gargione, Jimmy Tanamati Soares, and Prof. Jonathan Ruszala. Sponsors of LTU's Bigfoot team included DENSO, Realtime Technologies, Inc., Clearpath Robotics, Sparton, OmniSTAR, LTU's Department of Mathematics and Computer Science, the Barnes & Noble Bookstore at LTU, and Robofest.